\pdfoutput=1
\documentclass[11pt]{article}

\usepackage{tipa}
\usepackage[]{emnlp2021}

\usepackage{tikz} %important for graphs
\usetikzlibrary{arrows,automata,shapes,positioning} 
\usepackage{adjustbox}

\usepackage{times}
\usepackage{latexsym}

\usepackage{inconsolata}
\usepackage{amsmath}
\usepackage{amsfonts}
\usepackage{amsthm}
\usepackage{amssymb}

\usepackage{enumerate}
\usepackage{graphicx}
\usepackage{xspace}
\usepackage{hyperref}
\usepackage{booktabs}
\usepackage{thmtools} 
\usepackage{thm-restate}

\usepackage{cleveref}
\crefname{section}{\S}{\S\S}
\crefname{table}{Tab.}{}
\crefname{figure}{Fig.}{}
\crefname{algorithm}{Alg.}{}
\crefname{equation}{Eq.}{Eq.}
\crefname{appendix}{App.}{}
\crefname{theorem}{Theorem}{}
\crefname{prop}{Proposition}{}
\crefname{cor}{Corollary}{}
\crefname{observation}{Observation}{}
\crefname{assumption}{Assumption}{}
\crefname{hypothesis}{Hyp.}{Hypotheses}
\crefformat{section}{\S#2#1#3}

\newcommand{\inword}[1]{\textit{#1}}
\newcommand{\outword}[1]{\texttt{#1}}
\newcommand{\defn}[1]{\textbf{#1}}

\newtheorem{defin}{Definition}

\newtheorem{myexample}{Example}

\usepackage{todonotes}

\newcommand{\defeq}[0]{\mathrel{\stackrel{\textnormal{\tiny def}}{=}}}

\newcommand{\saveForCR}[1]

\everypar{\looseness=-1}

\usepackage{microtype}

\def\aclpaperid{3234} %  Enter the acl Paper ID here

\title{Benchmarking Compositionality with Formal Languages}

\usepackage{emoji}
\newcommand{\ethz}{\emoji[robot_emoji]{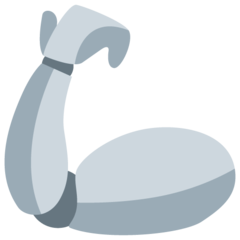}}
\newcommand{\ucambridge}{\emoji[robot_emoji]{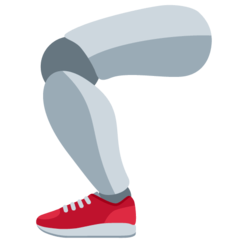}}
\newcommand{\nyu}{\emoji[robot_emoji]{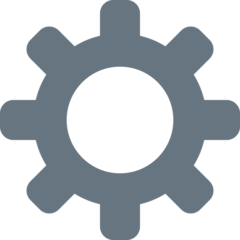}}
\newcommand{\sjsu}{\emoji[robot_emoji]{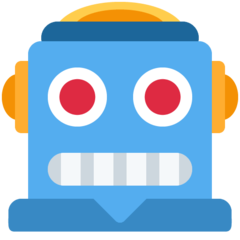}}
\newcommand{\fair}{\emoji[robot_emoji]{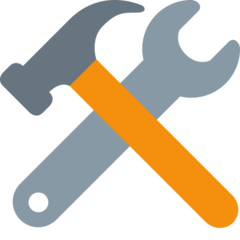}}

\author{Josef Valvoda$^{\ucambridge}$ Naomi Saphra$^{\nyu}$ Jonathan Rawski$^{\sjsu}$\\ 
\textbf{Adina Williams$^{\fair}$ Ryan Cotterell$^{\ethz}$}\\
$^{\ucambridge}$University of Cambridge
~\;~$^{\nyu}$New York University \\
~\;~$^{\sjsu}$San Jos\'e State University
~\;~$^{\fair}$FAIR
~\;~$^{\ethz}$ETH Z\"{u}rich \\
  \texttt{\href{mailto:jv406@cam.ac.uk}{jv406@cam.ac.uk}}~\;~ \texttt{\href{mailto:nsaphra@nyu.edu}{nsaphra@nyu.edu}}~\;~ \texttt{\href{mailto:jon.rawski@sjsu.edu}{jon.rawski@sjsu.edu}}\\
    \texttt{\href{mailto:adinawilliams@fb.com}{adinawilliams@fb.com}}~\;~
  \texttt{\href{mailto:ryan.cotterell@inf.ethz.ch}{ryan.cotterell@inf.ethz.ch}} \\ 
}

\date{}

\begin{document}

\maketitle
\begin{abstract}
Recombining known primitive concepts into larger novel combinations is a quintessentially human cognitive capability.
Whether large neural models in NLP can acquire this ability while learning from data is an open question.
In this paper, we investigate this problem from the perspective of formal languages.
We use deterministic finite-state transducers to make an unbounded number of datasets with controllable properties governing compositionality. By randomly sampling over many transducers, we explore which of their properties 
contribute to learnability of a compositional relation by a neural network. 
We find that the models either learn the relations completely or not at all. 
The key is transition coverage, setting a soft learnability limit at $400$ examples per transition.\looseness-1

\vspace{0.5em}
\hspace{.5em}\includegraphics[width=1.25em,height=1.25em]{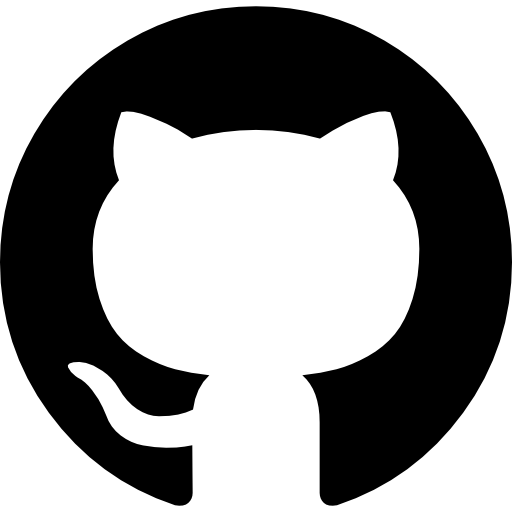}\hspace{.75em}\parbox{\dimexpr\linewidth-2\fboxsep-2\fboxrule}{\url{https://github.com/valvoda/neuralTransducer}}
\vspace{-.5em}
\end{abstract}

\section{Introduction}

Compositionality is a hallmark of human language 
\citep{montague-1970-universal, partee-1995-lexical, fodor-1998}.
It is arguably a requirement for any model to count as a model of language, or to achieve human-like natural language understanding.
Compositionality seems to be such a deep property of language that speakers draw conclusions about the overall meaning of sentences even when the meanings of individual words are not known. 
For instance, English speakers reading the Jabberwocky \cite{carroll-1871-through} comprehend that the noun phrase \inword{slithy toves} is built from the composition of the adjective \inword{slithy} with the plural noun \inword{toves}, despite lacking a clear understanding of what \inword{slithy} or \inword{toves}---let alone their composition---could mean.
In cognitive science, whether neural networks can learn to combine a limited number of primitives (in the case of language, word or morphemes) to describe a complex environment has been debated for over 30 years \cite{fodor-pylyshyn-1988-connectionism, marcus-1998-rethinking}.
\looseness=-1

In recent work, researchers have explored the inherent limitations of neural models to exhibit compositionality by analyzing sequence-to-sequence model performance on small, controlled datasets \citep{lake-baroni-2018-generalization, hewitt-etal-2020-rnns, ijcai2020-708, white-cotterell-2021-examining, dankers2021paradox,white+cotterell+coling2022}.
However, the conclusions of these studies are often murky. 
For instance, \citet{lake-baroni-2018-generalization} cast doubt on neural models' ability to do compositional generalization using their toy SCAN dataset, but shortly thereafter, \citet{bastings2018jump} demonstrated that an out-of-the-box sequence-to-sequence model could indeed fully master the task.
\looseness -1

Instead of hand-crafting small challenge datasets, we propose to test for compositionality by randomly sampling from a whole class of string-to-string functions.
In doing so, we draw on two linguistic traditions.
On the one hand, we follow Montague's assertion that no important theoretical difference exists between natural and artificial languages \cite{montague-1970-universal}.
Following this logic, the question of whether neural networks compositionally process human language is fundamentally equivalent to asking whether they compositionally process artificial languages.
On the other hand, we draw lessons from the field of grammatical inference \cite{de2010grammatical, rawskiheinz2019nfflml},
and evaluate neural sequence-to-sequence models on many automatically generated artificial languages sampled from particular classes of functions---as is standard practice at grammatical inference competitions \cite{pmlr-v57-balle16}.\looseness=-1

In this paper, we study the class of string functions encoded by subsequential finite-state transducers (SFSTs), a restricted class of general finite-state transducers \cite{mohri-1997-finite}.
We sample arbitrary SFSTs to generate many different string-to-string datasets and evaluate the behavior of neural sequence-to-sequence models when learning them.
By controlling the formal properties of the SFSTs we sample from, we are able to make precise statements about the learnability of systematic phenomena.\looseness=-1

Empirically, we find that neural sequence-to-sequence models are, in many cases, capable of perfectly learning SFSTs from finite data.
Moreover, we observe an interesting tendency for neural models to either generalize correctly or to fail outright---with little middle ground. 
Our analysis reveals a possible explanation for this---generalization seems to be possible when the training data has sufficient coverage, i.e., when every transition in a given transducer is crossed in a minimum number of training examples ($\approx 400$ in our experiments).

We then turn to analyze a popular hand-crafted dataset, SCAN, through the lens of an SFST that encodes it.
We find that SCAN is peculiar in that it seems to serve as a counterexample to our transition coverage finding.
This suggests that there is a more nuanced story to tell: 
We predict the learnability of a language based on a notion of complexity native to subregular languages, but it may be that a more consistently predictive complexity metric would come from a higher point on the Chomsky hierarchy. Future work might seek to limit the number of outlier languages like SCAN in order to identify a notion of complexity that is native to the architecture of the model itself. Such a notion of complexity would identify the level of abstraction that best reflects the representations learned by the model. A fruitful avenue towards this goal might lie in exploring more complex formalisms.
\looseness=-1

\section{Finite-State Transducers}
This section provides a short technical overview of finite-state transducers and motivates our choice to learn this class of relations.

\subsection{Why Learn Finite-State Transductions?}\label{sec:scan-fst}
Our study focuses on learning a particular kind of transduction. Specifically, we focus on restricted classes of regular relations, which are those relations computable by finite-state transducers.
We believe this is a natural starting point since this class of formal languages is mathematically well-studied, has provable learning guarantees, and has a long use history in linguistics and NLP \cite{mohri-1997-finite}.\looseness=-1

Finite-state transducers also encompass most previous work on compositionality: many datasets, e.g., SCAN \cite{lake-baroni-2018-generalization} and gSCAN \cite{ruis-etal-2020-benchmark}, describe \emph{finite} string relations and are, therefore, finite-state by definition.
These handcrafted datasets have many advantages, like easy interpretability and domain specificity since they directly encode particular relevant relationships like movement over a grid or specific linguistic phenomena. 
However, this realism pays the price of diminished robustness of any findings over such datasets \cite{RogersPullum11}, see \citet{white-cotterell-2021-examining} for more discussion of this point.
By removing the ability to simply adjust properties of the underlying function class, and the transducers which compute it, one loses the possibility to experiment more robustly over a function type, rather than just one token instantiation of it.
\looseness -1

Rather than manually designing individual datasets ourselves, we generate unboundedly many new datasets via randomly sampled SFSTs. This offers a principled view of the problem of learning artificial languages by simply varying properties of the class of transducers that generate them.
Furthermore, as we will see in \cref{sec:discussion}, one may view existing compositionality tasks as learning a specific SFST.
We contend this view enables a deeper understanding of modeling results.
Both specific artificial languages, such as the compositionality datasets mentioned above, and those randomly sampled from a particular function class such as the work presented in this paper, are worth studying. 
However, our approach has been missing from the compositionality discourse.

\subsection{Basic Theory}
Now we will overview the basic elements of finite-state theory that will be necessary for the rest of the paper; we start with some definitions.\looseness=-1

\newcommand{\mathcheck}[1]{#1}

\newcommand{\automaton}{\mathcheck{\mathcal{A}}}
\newcommand{\transducer}{\mathcheck{\mathcal{T}}}

\newcommand{\inputalphabet}{\mathcheck{\Sigma}}
\newcommand{\outputalphabet}{\mathcheck{\Gamma}}
\newcommand{\states}{\mathcheck{Q}}
\newcommand{\initialstate}{\mathcheck{q_0}}
\newcommand{\finalstates}{\mathcheck{F}}
\newcommand{\transitions}{\mathcheck{\delta}}

\newcommand{\transtuple}{\mathcheck{\left\langle \inputalphabet, \outputalphabet, \states, \initialstate, \finalstates, \transitions \right\rangle}}
\newcommand{\autotuple}{\mathcheck{\left\langle \inputalphabet, \states, \initialstate, \finalstates, \transitions \right\rangle}}

\newcommand{\str}{\mathcheck{\boldsymbol{\sigma}}}
\newcommand{\strout}{\mathcheck{\boldsymbol{\gamma}}}
\newcommand{\kleene}[1]{#1^*}
\newcommand{\zerostring}{\mathcheck{\boldsymbol{0}}} %zerostring

\begin{defin}
A \defn{finite-state automaton} (FSA) is a 5-tuple $\automaton=\autotuple$ where
\begin{itemize}
    \setlength\itemsep{0.1em}
    \item $\inputalphabet$ is an input alphabet whose elements are denoted $\sigma$;
    \item $\states$ is a finite set of states whose elements are denoted $q$;
    \item $\initialstate \in \states$ is the unique start state;
    \item $\finalstates \subseteq \states$ is the set of final states;
    \item $\transitions: \states \times \inputalphabet \cup \{ \varepsilon \} \rightarrow \states$ is the transition relation and $\varepsilon$ is an empty string.
\end{itemize}
\noindent We denote transitions, i.e., when $q' \in \transitions(q, \sigma)$, with the more suggestive notation $q \xrightarrow[]{\sigma} q'$.
We say that the automaton $\automaton$ \defn{accepts} a string $\str \in \kleene{\inputalphabet}$ iff there exists a path\footnote{A path is a sequence of transitions.} through the automaton states $\initialstate \xrightarrow[]{\sigma_1} q_1 \xrightarrow[]{\sigma_2} q_2  \cdots \xrightarrow[]{\sigma_N} q_N$ where $\initialstate$ is the initial state and $q_N \in \finalstates$ and $\sigma_1 \cdots \sigma_N  = \str$.
In our notation $\sigma_n$ can be the empty string $\varepsilon$.
We call $\str$ the \defn{yield} of the path $\initialstate \xrightarrow[]{\sigma_1} q_1 \xrightarrow[]{\sigma_2} q_2  \cdots \xrightarrow[]{\sigma_N} q_N$.
\end{defin}

Furthermore, we call an automaton \textbf{complete} when one may transition from every state to every symbol, i.e., $\transitions(q, \sigma)$ is defined for all $q \in \states$ and $\sigma \in \inputalphabet$.
And, we say an automaton is \textbf{deterministic}, if, given a state $q$ and an alphabet symbol $\sigma \in \inputalphabet$, there is at most one transition for $\sigma$ from $q$, i.e., 
$|\transitions(q, \sigma)| \leq 1$.
We have that $|\transitions(q, \sigma)| = 1$ for all $q \in \states$ and $\sigma \in \inputalphabet$ if the automaton is \emph{both} complete and deterministic.
In our examples below we denote the \textbf{final state} as a circle with a double border.\looseness-1
 
\begin{myexample}
Below we exhibit a complete deterministic finite-state automaton.
\begin{center}
\begin{tikzpicture}[scale=0.19, every node/.style={scale=0.9}]
\tikzstyle{every node}+=[inner sep=0pt]
\filldraw[black, fill=lightgray!40, thick]  (41.7,-20.5) circle (3);
\draw (41.7,-20.5) node {$q_0$};
\filldraw[black, fill=lightgray!40, thick]  (52.6,-20.5) circle (3);
\draw (52.6,-20.5) node {$q_1$};
\draw [black] (52.6,-20.5) circle (2.4);
\draw [black] (43.613,-18.229) arc (125.70549:54.29451:6.06);
\fill [black] (50.69,-18.23) -- (50.33,-17.36) -- (49.75,-18.17);
\draw (47.15,-16.59) node [above] {$a$};
\draw [black] (51.017,-23.008) arc (-47.33205:-132.66795:5.705);
\fill [black] (43.28,-23.01) -- (43.53,-23.92) -- (44.21,-23.18);
\draw (47.15,-25.02) node [below] {$b$};
\draw [black] (39.02,-21.823) arc (324:36:2.25);
\draw (34.45,-20.5) node [left] {$b$};
\fill [black] (39.02,-19.18) -- (38.67,-18.3) -- (38.08,-19.11);
\draw [black] (55.28,-19.177) arc (144:-144:2.25);
\draw (59.85,-20.5) node [right] {$a$};
\fill [black] (55.28,-21.82) -- (55.63,-22.7) -- (56.22,-21.89);
\end{tikzpicture}
\end{center}
\noindent The above finite-state automaton accepts the language $\{(b^{i}a^{j})^n \mid i,j,n \in \mathbb{Z}_{+}\}$.
\end{myexample}

\begin{defin}
A \defn{finite-state transducer} (FST) is a 6-tuple $\transducer=\transtuple$:
\begin{itemize}
    \setlength\itemsep{0.1em}
    \item $\inputalphabet$ is an input alphabet whose elements are denoted $\sigma$;
     \item $\outputalphabet$ is an output alphabet whose elements are denoted $\gamma$;
    \item $\states$ is a finite set of states whose elements are denoted $q$;
    \item $\initialstate \in \states$ is the unique start state;
    \item $\finalstates \subseteq \states$ is the set of final states;
    \item $\transitions: \states \times \inputalphabet \cup \{ \varepsilon \} \rightarrow \states \times \kleene{\outputalphabet}$ is the transition relation.
\end{itemize}
We denote transitions, i.e., when $(q',\gamma) \in \transitions(q, \sigma)$, with the more suggestive notation $q \xrightarrow[]{\sigma / \strout} q'$.
We say that the transducer $\transducer$ \defn{transduces} a string $\str \in \kleene{\inputalphabet}$ to a string $\strout \in \kleene{\outputalphabet}$ iff there exists a path $\initialstate \xrightarrow[]{\sigma_1 / \strout_1} q_1 \xrightarrow[]{\sigma_2 / \strout_2} q_2  \cdots \xrightarrow[]{\sigma_N / \strout_N} q_N$ where $\initialstate$ is the initial state and $q_N \in \finalstates$, $\sigma_1 \cdots \sigma_N  = \str$
and $\strout_1 \cdots \strout_N = \strout$.
In our notation either $\sigma_n$ or $\strout_n$ can be the empty string $\varepsilon$, from which it follows that the length $|\str|$ must not equal $|\strout|$. 
We call $\str$ the \defn{input yield} and $\strout$ the \defn{output yield}, respectively, of the path $\initialstate \xrightarrow[]{\sigma_1 / \strout_1} q_1 \xrightarrow[]{\sigma_2 / \strout_2} q_2  \cdots \xrightarrow[]{\sigma_N / \strout_N} q_N$.
\end{defin}

As in the case of an FSA, we say an FST is \defn{complete} if $\transitions$ is defined for all states and all symbols, i.e., $|\transitions(q, \sigma)| > 0$ for $q \in \states$ and $\sigma \in \inputalphabet$.
Furthermore, we say an FST is \defn{subsequential} if it is deterministic with respect to the input, i.e., if $|\transitions(q, \sigma)| \leq 1$ for all $q \in \states$ and $\sigma \in \inputalphabet$ and the FST does not have transitions of the form $q \xrightarrow[]{\varepsilon / \strout} q'$.
By construction, subsequential transducers (SFSTs) are \defn{functional}, i.e., the string-to-string relations they encode are functions rather than relations.
Indeed, it is this functionality that makes them a useful tool for the analysis of neural sequence-to-sequence models.
The class of \defn{subsequential functions} are those describable with SFSTs.\footnote{Note there are other algebraic and logical characterisations \citep{oncina-etal-1993-learning,bhaskaretal20}.}
They are a subclass of the regular relations, but a superclass of the finite relations.\looseness=-1

\begin{myexample}\label{example:fst}
Below is an example of a non-deterministic finite-state transducer:
\begin{center}
\begin{tikzpicture}[scale=0.19, every node/.style={scale=0.9}]
\tikzstyle{every node}+=[inner sep=0pt]
\filldraw[black, fill=lightgray!40, thick] (27.9,-20.5) circle (3);
\draw (27.9,-20.5) node {$q_0$};
\filldraw[black, fill=lightgray!40, thick] (40.5,-20.5) circle (3);
\draw (40.5,-20.5) node {$q_1$};
\filldraw[black, fill=lightgray!40, thick] (53.5,-20.5) circle (3);
\draw (53.5,-20.5) node {$q_2$};
\draw [black] (53.5,-20.5) circle (2.4);
\draw [black] (30.9,-20.5) -- (37.5,-20.5);
\fill [black] (37.5,-20.5) -- (36.7,-20) -- (36.7,-21);
\draw (34.2,-21) node [below] {$a/b$};
\draw [black] (43.5,-20.5) -- (50.5,-20.5);
\fill [black] (50.5,-20.5) -- (49.7,-20) -- (49.7,-21);
\draw (47,-21) node [below] {$b/\varepsilon$};
\draw [black] (30.303,-18.709) arc (122.28226:57.71774:19.467);
\fill [black] (51.1,-18.71) -- (50.69,-17.86) -- (50.15,-18.7);
\draw (40.7,-15.2) node [above] {$\varepsilon/b$};
\end{tikzpicture}
\end{center}
The above transducer only has two paths $\initialstate \xrightarrow[]{a / b} q_1 \xrightarrow[]{b / \varepsilon} q_2$ and $\initialstate \xrightarrow[]{\varepsilon / b} q_2$. 
The first transduces $ab \mapsto b$ and the second $\varepsilon \mapsto b$.\looseness=-1
\end{myexample}
% \ryan{Not sure the [] labels are helping here...}
\begin{myexample}\label{example:sfst}
% [A Complete Subsequential Finite-State Transducer]\label{example:sfst}
Below is an example of a complete subsequential transducer over the input alphabet $\inputalphabet = \{a, b\}$ and output alphabet $\outputalphabet = \{a, b\}$.
\begin{center}
\begin{tikzpicture}[scale=0.19, every node/.style={scale=0.9}]
\tikzstyle{every node}+=[inner sep=0pt]
\filldraw[black, fill=lightgray!40, thick] (22.4,-25.5) circle (3);
\draw (22.4,-25.5) node {$q_0$};
\filldraw[black, fill=lightgray!40, thick] (33.9,-25.5) circle (3);
\draw (33.9,-25.5) node {$q_1$};
\draw [black] (33.9,-25.5) circle (2.4);
\draw [black] (19.72,-26.823) arc (-36:-324:2.25);
\draw (15.15,-25.5) node [left] {$b/b$};
\fill [black] (19.72,-24.18) -- (19.37,-23.3) -- (18.78,-24.11);
\draw [black] (24.434,-23.329) arc (123.94508:56.05492:6.655);
\fill [black] (31.87,-23.33) -- (31.48,-22.47) -- (30.92,-23.3);
\draw (28.15,-21.7) node [above] {$a/\varepsilon$};
\draw [black] (32.229,-27.955) arc (-48.26855:-131.73145:6.128);
\fill [black] (24.07,-27.96) -- (24.34,-28.86) -- (25,-28.11);
\draw (28.15,-30.01) node [below] {$b/\varepsilon$};
\draw [black] (36.58,-24.177) arc (144:-144:2.25);
\draw (41.15,-25.5) node [right] {$a/a$};
\fill [black] (36.58,-26.82) -- (36.93,-27.7) -- (37.52,-26.89);
\end{tikzpicture}
\end{center}
Note the absence of $\varepsilon$ on the input side; however, we do have $\varepsilon$ on the output side.
\end{myexample}

\section{Compositionality Formalized}\label{sec:formalized}
It is widely held that compositionality is a cornerstone of human language.
However, language researchers use the term compositionality to refer to a variety of different concepts \cite{decomposed}.
Here, we discuss the specific definition we employ throughout the paper and give a theoretical justification for the use of finite-state transducers as an instantiation of that definition.\looseness=-1
\subsection{Montague's Compositionality\footnote{\color{teal} 
\textbf{A clarifying note}: Some readers, post-publication, reported that they interpreted our paper as making a general claim about the \emph{natural language} syntax--semantics interface.
Specifically, that it can be well modeled as a regular string-to-string transduction.
This was \emph{not} our intent.
Instead, our goal is to exhibit a simple benchmark based on SFSTs that is formally compatible with Montague's definition of compositionality and allows for extreme experimentation.\looseness=-1
}}

Montague famously gave a mathematical definition of what it means for a language to be compositional \cite{montague-1970-universal}---specifically, he proposed that a relation, e.g., the mapping from syntax to semantics, be called compositional if and only if it is a homomorphism, i.e., if it is a structure-preserving map between an input and an output algebra \cite{andreas}.
%In the context of string-to-string maps, a function that preserves concatenation is a homomorphism.
To make this notion more formal, we have to be specific about what structure we hope our function preserves.
In this paper, we will exclusively focus on monoidal structure.
We emphasize, however, that the notion of a homomorphism is not restricted to monoids.
\begin{defin}
A \defn{monoid} is a set $A$ equipped with a binary operation $\bullet$ such that\looseness=-1
\begin{itemize}
\setlength\itemsep{0.1em}
\item For $f, g\in A$, $f \bullet g \in A$ (\defn{closure});
\item For $f,g,h \in A$, $f \bullet (g \bullet h) = (f \bullet g) \bullet h$ (\defn{associativity});
\item There exists a unique element $e$ such that for every $g \in A$, we have that $g \bullet e = e \bullet g = g$ (\defn{identity}). 
\end{itemize}
\end{defin}
\begin{defin}
A \defn{free monoid} over strings is the structure $(\kleene{\inputalphabet}, \circ)$ where $\kleene{\inputalphabet}$ is the Kleene closure of the alphabet $\inputalphabet$ and $\circ$ is string concatenation.\footnote{When clear from context, we write $\sigma\circ\sigma'$ as $\sigma\sigma'$.\looseness=-1}
The empty string $\varepsilon$ is the identity element as $\varepsilon \circ \str = \str \circ \varepsilon = \str$ for any $\str \in \kleene{\inputalphabet}$.
\end{defin}

\begin{defin}
Let $(A, \bullet_A)$ and $(B, \bullet_B)$ be two 
monoids with unique identity elements $e_A$ and $e_B$.
We call a function $f : A \rightarrow B$ a \defn{homomorphism} if it obeys the following two properties:
\begin{itemize}
    \itemsep0em
    \item $f(e_A) = e_B$;
    \item $f(x \bullet_A y) = f(x) \bullet_B f(y), \quad \forall x, y \in A$
\end{itemize}
\end{defin}

We call a homomorphism between two free monoids a \defn{string homomorphism}.
As it turns out, there is a precise connection between string homomorphisms and finite-state theory.

\begin{restatable*}{proposition}{singlestate}\label{prop:single-state}
Let $\inputalphabet$ and $\outputalphabet$ be two alphabets.
The function $f$ is a string homomorphism between $(\kleene{\inputalphabet}, \circ)$ and $(\kleene{\outputalphabet}, \circ)$ iff it is a minimal non-empty complete subsequential finite-state transducer with one state.\looseness=-1
\end{restatable*}
\begin{proof}
See \Cref{appendix}.
\end{proof}
\Cref{prop:single-state} starts to shed light on the connection between Montague's notion of compositionality and finite-state transducers. 
However, this connection is quite weak because multi-state transducers are not covered.
We remedy this disparity in the subsequent section.

\subsection{Transducers as Homomorphisms}\label{sec:sfst}

Now we offer a more formal treatment of the exact sense in which SFSTs may be considered homomorphism and thus fall under Montague's definition of compositionality.
As shown by \Cref{prop:single-state}, in general, SFSTs do \emph{not} encode string homomorphisms.
Indeed, it is straightforward to find a counterexample that hammers this point home.
\begin{myexample}\label{example:no-homo}
Below we exhibit a two-state subsequential finite-state transducer that is \emph{not} a string homomorphism.\looseness=-1
\vspace{.1cm}
\begin{center}
\begin{tikzpicture}[scale=0.19, every node/.style={scale=0.9}]
\tikzstyle{every node}+=[inner sep=0pt]
\filldraw[black, fill=lightgray!40, thick] (27.9,-20.5) circle (3);
\draw (27.9,-20.5) node {$q_0$};
\filldraw[black, fill=lightgray!40, thick] (40.5,-20.5) circle (3);
\draw (40.5,-20.5) node {$q_1$};
\draw [black] (40.5,-20.55) circle (2.4);
\filldraw[black, fill=lightgray!40, thick] (53.5,-20.5) circle (3);
\draw (53.5,-20.5) node {$q_2$};
\draw [black] (53.5,-20.5) circle (2.4);
\draw [black] (30.9,-20.5) -- (37.5,-20.5);
\fill [black] (37.5,-20.5) -- (36.7,-20) -- (36.7,-21);
\draw (34.2,-21) node [below] {$a/a$};
\draw [black] (43.5,-20.5) -- (50.5,-20.5);
\fill [black] (50.5,-20.5) -- (49.7,-20) -- (49.7,-21);
\draw (47,-21) node [below] {$b/b$};
\draw [black] (30.303,-18.709) arc (122.28226:57.71774:19.467);
\fill [black] (51.1,-18.71) -- (50.69,-17.86) -- (50.15,-18.7);
\draw (40.7,-15.2) node [above] {$b/a$};
\end{tikzpicture}
\end{center}
In the above example, we have $ab \mapsto ab$, but also $a \mapsto a$ and $b \mapsto a$.
Thus, it is not a homomorphism.
\end{myexample}

\newcommand{\matrixA}{\boldsymbol{A}}
\newcommand{\matrixB}{\boldsymbol{B}}
\newcommand{\matrixSigma}{\boldsymbol{M}^{\sigma}}
\newcommand{\matrixEps}{\boldsymbol{M}^{\epsilon}}
\newcommand{\matrixSigmaPrime}{\boldsymbol{M}^{\sigma'}}
\newcommand{\matrixentrySigma}{M^\sigma}
\newcommand{\matrixentrySigmaPrime}{M^{\sigma '}}

\Cref{example:no-homo} is dissatisfying; it contradicts the intuition that an SFST encodes some notion of Montague-esque compositionality.  Luckily, as it turns out, we can find a precise sense in which an SFST is indeed a homomorphism. 
The idea is to lift the free monoid into a matrix.
Given a complete SFSA $\automaton = \autotuple$ over $|\states| = N$ states, for every symbol $\sigma \in \inputalphabet$ define an $N 
\times N$ \defn{symbol transition matrix} $\matrixSigma$ whose entries are \looseness=-1
\begin{equation}
   \matrixentrySigma_{ik} \defeq
\begin{cases}
    \sigma,& \textbf{if } q_i \xrightarrow[]{\sigma} q_k \in \transitions\\
    \zerostring,              & \textbf{otherwise}
\end{cases}
\end{equation}
where $\zerostring$ is a distinguished symbol, which is not in $\inputalphabet$, called the \defn{zero string}.
The zero string is an \defn{anhilitator}, i.e., it has the property that $\zerostring \circ \sigma = \sigma \circ \zerostring = \zerostring$ for any $\sigma \in \inputalphabet$.

Importantly, because $\automaton$ is complete and deterministic, there is exactly one non-$\zerostring$ entry in every row of $\matrixSigma$. 
Now, we define an operation $\otimes$ between two such matrices.
Since the automaton is complete and deterministic, we know there exists a unique $j'$ such that $\matrixentrySigma_{ij'} \neq \zerostring$ and, by the same argument, there exists a unique $k'$ such that $\matrixentrySigma_{j'k'} \neq \zerostring$.
Then, in terms of $j'$ and $k'$, we have
\begin{equation}\label{eq:matrix-multiplication}
    (\boldsymbol{M}^{\sigma} \otimes \boldsymbol{M}^{\sigma'})_{ik} = 
    \begin{cases}
        M^{\sigma}_{ij'}\circ M^{\sigma}_{j' k'}, & \textbf{if } k = k'\\
        \zerostring, & \textbf{otherwise}
    \end{cases}
\end{equation}
Clearly, $\boldsymbol{M}^{\sigma} \otimes \boldsymbol{M}^{\sigma'}$, for any $\sigma, \sigma' \in \Sigma$,
enforces that the resulting product still has the property that there is exactly one element of every row that is not equal to $\zerostring$.\looseness=-1
\footnote{This property is reminiscent of the ``tails'' of a subsequential function \cite{oncina-etal-1993-learning}}
With $\boldsymbol{M}^{\boldsymbol{\sigma}}$, where $\boldsymbol{\sigma} = \sigma_1 \cdots \sigma_K \in \Sigma^*$ is a string of length $K$, we denote the product of matrices $\boldsymbol{M}^{\sigma_1} \otimes \cdots \otimes \boldsymbol{M}^{\sigma_K}$.

\begin{restatable*}{proposition}{monoid}
Let $\automaton = \autotuple$ be a complete deterministic finite-state automaton.
Let $\mathcal{M} \defeq \Big\{ \matrixSigma \mid \sigma \in \inputalphabet\Big\}$ be the set of $\automaton$'s symbol transition matrices. 
Then $(\mathcal{M}^*, \otimes)$ with $\otimes$ defined in \cref{eq:matrix-multiplication} is a \defn{transition monoid} with 
$\boldsymbol{E}$ as a distinguished identity element.\footnote{The element $\boldsymbol{E}$ can be thought of as an $N \times N$ matrix where every element is $\varepsilon$.}\looseness=-1
\end{restatable*}
\begin{proof}
See \Cref{appendix}.
\end{proof}

Now we are in a position to discuss the precise sense in which subsequential finite-state transducers are homomorphisms.
In the case of a finite-state transducer $\transducer = \transtuple$, there are two
symbol transition matrices.
The \emph{input} symbol transition matrix is defined analogously to that of a finite-state acceptor.
However, the \emph{output} symbol transition matrix is defined slightly differently:
\begin{equation}
   \matrixentrySigma_{ik} \defeq 
\begin{cases}
    \strout,& \textbf{if } q_i \xrightarrow[]{\sigma / \strout} q_k \in \transitions\\
    \zerostring,              & \textbf{otherwise}
\end{cases}
\end{equation}
Indeed, for any product of $K$ output symbol transition matrices $\boldsymbol{M}^{\boldsymbol{\sigma}} = \boldsymbol{M}^{\sigma_1} \otimes \cdots \otimes \boldsymbol{M}^{\sigma_K}$, the entries of 
$\boldsymbol{M}^{\boldsymbol{\sigma}}$ have a clear interpretation.
Let $\str = \sigma_1 \cdots \sigma_K$ and suppose
$\boldsymbol{M}^{\boldsymbol{\sigma}}_{ik} = \strout \neq \zerostring$.
Then, we know that if we start in state $q_i$ and read in input string $\str$ we end up in state $q_k$ and output string $\strout$. 
There are exactly $|\states|$ non-zero entries in $\boldsymbol{M}^{\boldsymbol{\sigma}} $.
Thus, for any given finite-state transducer there is both an input and output transition monoid, which we denote as $(\mathcal{M}^*_\inputalphabet, \otimes)$ and $(\mathcal{M}^*_\outputalphabet, \otimes)$, respectively.
Then, the intuition is that SFSTs constitute a homomorphism over the closure of the symbol transition matrices. 
We state the more formal result below.

\begin{restatable*}{proposition}{matrix}\label{prop:matrix-compositional}
Let $\transducer = \transtuple$ be a complete subsequential finite-state transducer. 
Let $(\mathcal{M}^*_\inputalphabet, \otimes)$ be the transition monoid associated with the input alphabet $\inputalphabet$ and let $(\mathcal{M}^*_\outputalphabet, \otimes)$ be the transition monoid associated with the output alphabet $\outputalphabet$.
Then there exists a homomorphism $f : \mathcal{M}^*_\inputalphabet \rightarrow \mathcal{M}^*_\outputalphabet$.
\end{restatable*}
\begin{proof}
See \Cref{appendix}.
\end{proof}
At a higher level, \Cref{prop:matrix-compositional} simply fixes the bug present in \Cref{example:no-homo} by incorporating the state into the values.
Moreover, \Cref{prop:matrix-compositional} is a clear generalization of \Cref{prop:single-state} in that if we have a single-state SFST, we have $1\times 1$ matrices that may be viewed as single symbols and the operation $\otimes$ defined in \Cref{eq:matrix-multiplication} reduces to string concatenation.

\paragraph{Briefly back to Montague.}
To put the above results in context, we showed by \Cref{prop:matrix-compositional} that arbitrary complete SFSTs 
are compositional in the sense of Montague.
Specifically, SFSTs encode a homomorphism between the free monoid $\left(\kleene{\inputalphabet}, \cdot\right)$, which is isomorphic to $\left(\mathcal{M}_{\inputalphabet}^*, \otimes \right)$, and the transition monoid $\left(\mathcal{M}_{\outputalphabet}^*, \otimes \right)$.
This instantiates compositionality by directly linking it to the robust class-based characteristics of subsequential functions. There are of course other ways, but subsequentiality gives us a theoretical underpinning to use learning SFSTs from finite data as a benchmark for compositionality, and motivates our experiments and analysis in the coming sections. \looseness=-1

\section{Experimental Methods}
Next we introduce the methodology for generating our datasets as well as the neural and symbolic models employed in the empirical portion of our paper.\looseness=-1

\subsection{Generating Random SFSTs}\label{sec:sampling-sfsts}
We generate random SFSTs using the following stochastic process.
We first generate random unlabeled directed graphs that correspond to the symbol-specific transition matrices.
Given a set of states $Q$ (let $N = |Q|$) and an input alphabet $\inputalphabet$, we sample a matrix $\matrixB^{\sigma} \in \mathbb{B}^{N \times N}$ for every $\sigma \in \Sigma$ where $\mathbb{B} = \{0, 1\}$.
During sampling, we enforce the constraint that there be \emph{at most} one non-zero entry in every row vector $\mathbf{b}^{\sigma}_{i}$.
This constraint ensures that the resulting SFST is subsequential by construction.
To achieve this, we sample from $\{1, \ldots, N\}$ uniformly at random for each of the $N$ rows to determine the location of the non-zero entry in each of the $N$ rows.
In terms of the SFST, if the entry $b^{\sigma}_{ik} = 1$, then our generated SFST has a transition from state $q_i \overset{\sigma}{\longrightarrow} q_k$ with input symbol $\sigma$.
Then, for every transition $q_i \overset{\sigma}{\longrightarrow} q_k$ in our generated SFST, we sample its output symbol $\gamma$ from a uniform distribution over the output alphabet $\Gamma$.
This results in a transition $q_i \overset{\sigma / \strout}{\longrightarrow} q_k$.
Finally, to get a canonical representation
for particular SFSTs, we perform finite-state minimization \cite{CHOFFRUT2003131}.
Minimization also ensures that our sampled SFSTs are canonicalized SFSTs which allows us to check duplicates and ensure the same SFST is not sampled more than once.\looseness=-1

\subsection{Generating the Datasets}
Next we discuss the creation of the datasets we use to investigate the learnability of our sampled SFSTs.
We start by sampling $1000$ unique SFSTs using the process described in \cref{sec:sampling-sfsts}. 
All SFSTs have an input alphabet $\inputalphabet$ of 10 symbols and output alphabet of 30 symbols. 
We consider SFSTs with states numbering from $10$ to $100$ in increments of $10$, and we sample $100$ unique SFSTs for each number. 
Additionally, we sample extra SFSTs with $21$ to $39$ states because it was at this point that we empirically observed performance drop-off during a preliminary investigation. 
In total, our experiments use $2800$ unique SFSTs.

\paragraph{Sampling Input--output Pairs.}
To generate the input--output pairs to train and evaluate a neural model, we perform a random walk through the SFST where we select a transition (including the option to halt if we are in a final state) uniformly at random. 
Following \citet{lake-baroni-2018-generalization}, the maximum length of an input string is capped at $50$, i.e., we reject walks with more than $50$ steps during sampling. 
Using this process, we collect $20,000$ unique input--output pairs from each SFST.
We considered larger dataset sizes ($40,000$) for the SFSTs with $29$ to $39$ states since we observe an accuracy drop-off in this region.
All datasets are randomly split $80$--$20$ into training and test sets.

\begin{figure}
    \centering
    \includegraphics[width=\columnwidth, trim=0 0 0 0 0]{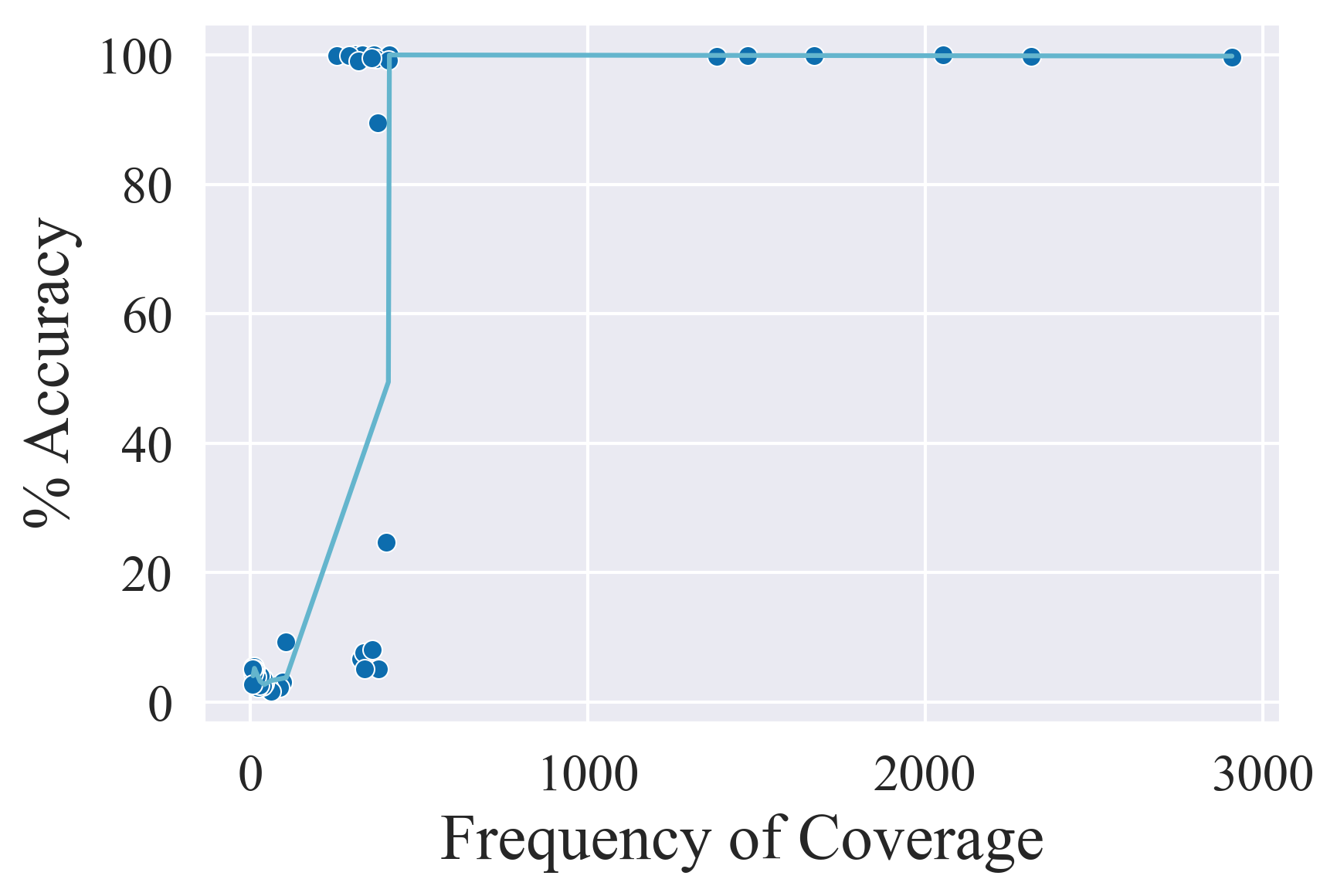}
    \setlength{\belowcaptionskip}{-5pt}
    \caption{Accuracy versus frequency of transition coverage, with an inflection point at $400$.}
    \label{fig:freq}
    \vspace{-10pt}
\end{figure}

\subsection{Neural Sequence-to-Sequence Models}\label{sec:seq2seq}

Our experiments make use of \citeposs{wu-cotterell-2019-exact} open-source neural transduction library.\footnote{The code is available at \url{https://github.com/shijie-wu/neural-transducer}.}
Our experiments consider an LSTM encoder--decoder with attention in the style of \newcite{BahdanauCB14}.
We use the following hyperparameters: $200$-dimensional hidden states in the encoder and decoder, each of which have $2$ layers, maximum gradient clipping normalization of $5$, dropout set to $0.5$, and a batch size of $64$.
Additionally, our alphabet tokens are embedded as $100$-dimensional vectors.
We train the model for $100,000$ epochs using the Adam optimizer \cite{KingmaB14} with the default learning rate of $0.001$.
To determine the effect of model capacity, we also consider a neural sequence-to-sequence model with $300$-dimensional hidden states and all other hyperparameters kept the same.\looseness=-1

\subsection{OSTIA}\label{sec:ostia}

The onward subsequential transducer inference algorithm \citep[OSTIA;][]{oncina-etal-1993-learning} learns the class of subsequential functions from positive presentations of input--output strings. 
OSTIA works by first building a prefix-tree transducer of the training data, which is then transformed through a series of state-merging operations into the SFST encoding the function the data is drawn from. If a characteristic sample is contained in the learning data, OSTIA finds a correct transducer in polynomial (cubic) time. 
Since OSTIA is designed specifically to learn subsequential relations, it provides a useful baseline. Unfortunately, due to its cubic runtime, OSTIA is too slow to use on larger datasets.
To give a practical speed-up, we limit the number of samples we provide to OSTIA to 1000, which is only 5\% to 10\% of what the neural transducers train on.
This keeps OSTIA's run time roughly equivalent to its neural counterparts, but unlevels the playing field, as it were.\looseness=-1

\begin{figure}[t]
    \centering
    \includegraphics[width=\columnwidth, trim=0 0 0 0 0]{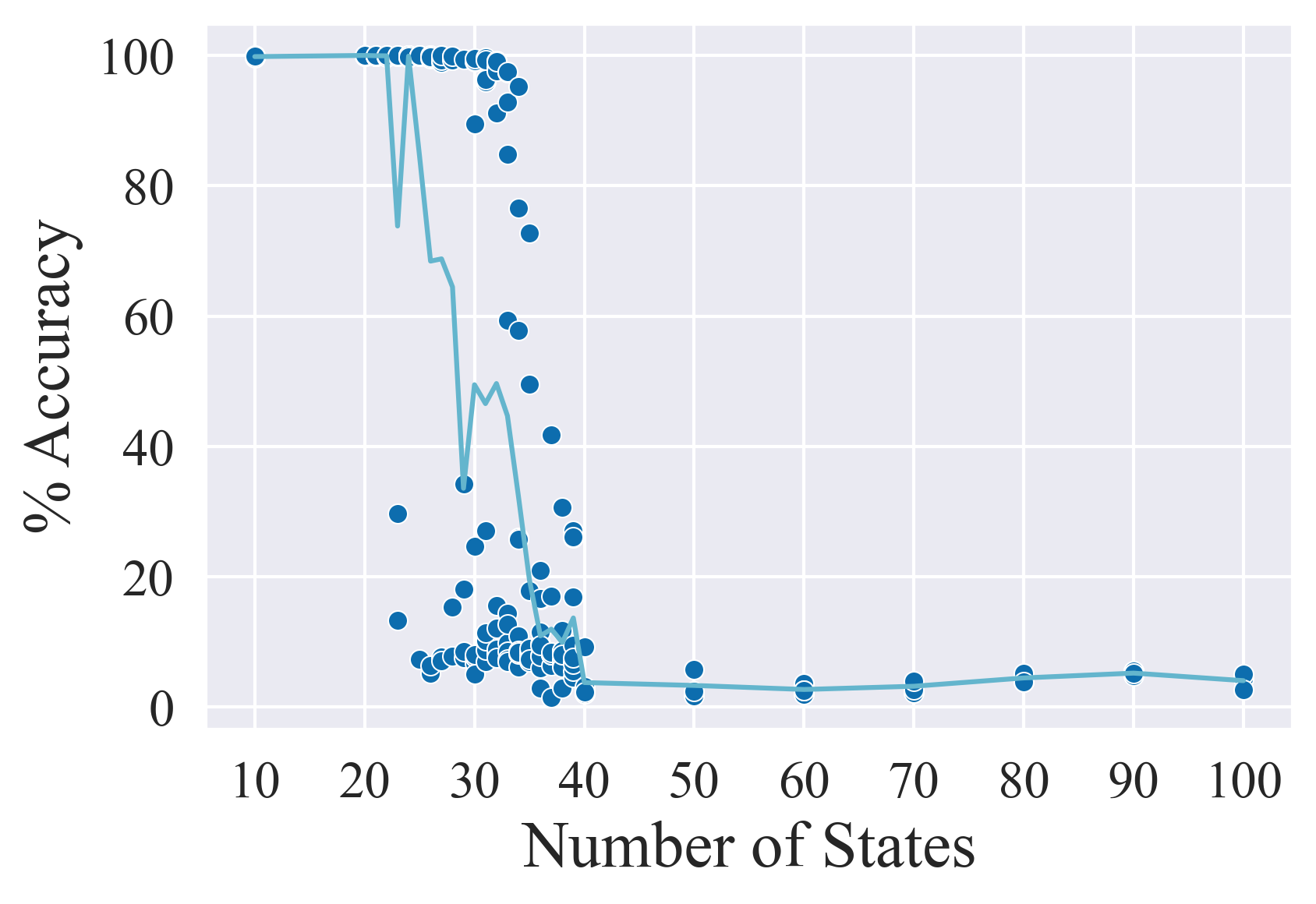}
    \caption{As the number of states in SFST increases, accuracy drops to nearly zero.}
    \label{fig:acc_dropoff}
    \vspace{-15pt}
\end{figure}

\section{Results and Discussion}\label{sec:results}
We train a neural network, as described in \cref{sec:seq2seq}, on the datasets taken from our sampled SFSTs. 
We discuss and analyze the results below.\looseness=-1

\paragraph{Minimum Transition Coverage.}
We define transition coverage of a given transition as the percentage of samples in the training dataset that cross that transition.
The results reported in \cref{fig:freq} reveal a threshold on the number of samples per transition required to comfortably learn the transducer: $400$ samples. 
This may seem unsurprising given neural networks' ``notorious thirst'' for data \citep{lake-baroni-2018-generalization}. 
In the vast majority of the cases when we train on a dataset that does not meet our transition coverage threshold, the neural network does not generalize to held-out data.
And, indeed, in the few cases where they manage to have non-zero accuracy, we observe that early transitions in such SFSTs have attained sufficient coverage, and are responsible for the above zero performance.
These findings give credence to the idea that there is a relatively simple complexity metric on the SFSTs, i.e., transition coverage, that determines whether or not the neural model will generalize.

\begin{figure}
    \centering
    \includegraphics[width=\columnwidth, trim=0 0 0 0 0]{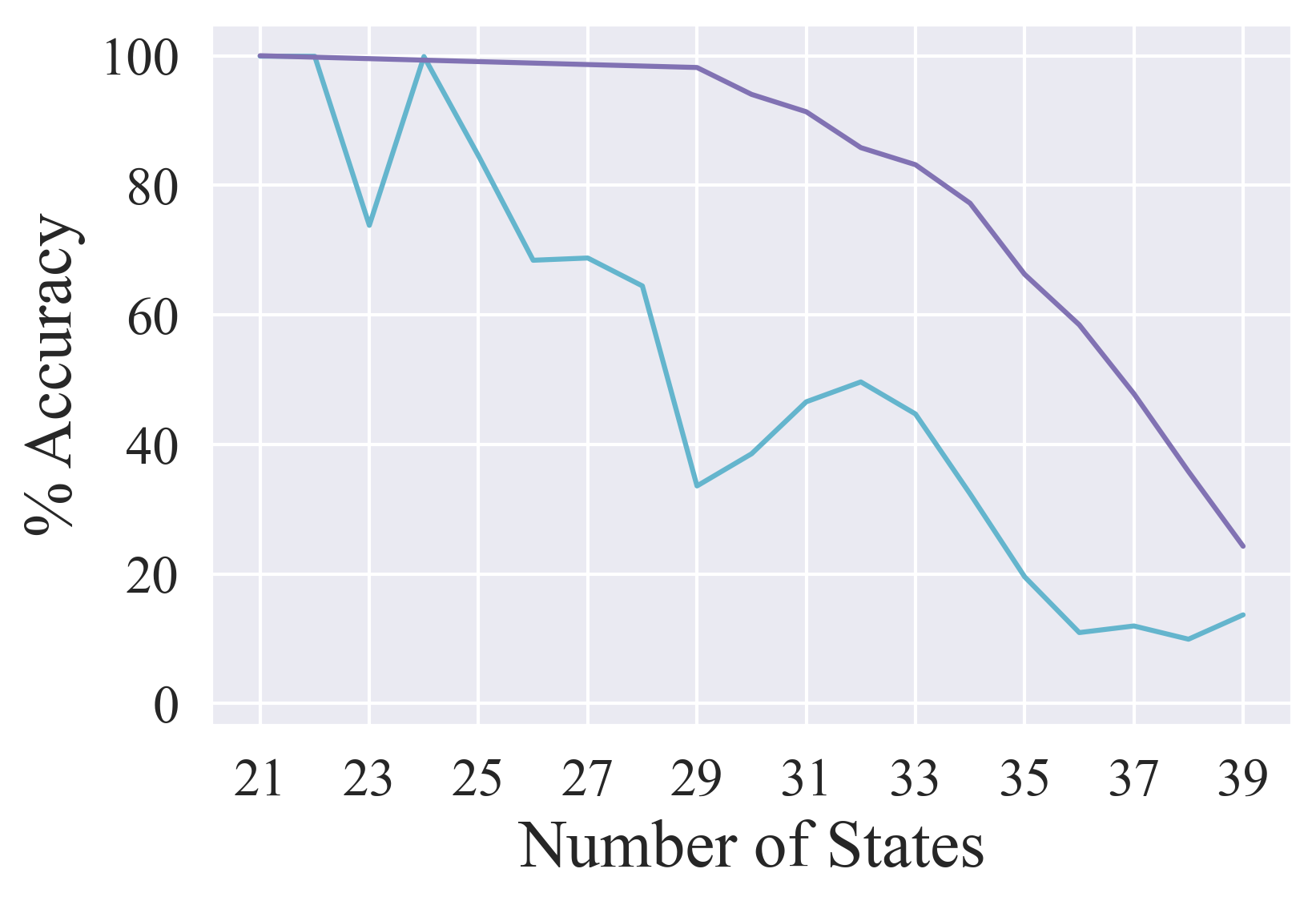}
    \setlength{\belowcaptionskip}{-15pt}
    \caption{As SFST state size increases, average accuracy (blue line) decreases. Increasing the neural model's size improves accuracy (purple line). 
    }
    \label{fig:20_acc_dropoff}
\end{figure}

\paragraph{Bigger Models Generalize Better.}
Additionally, we find that our discovered transition coverage threshold is \emph{not} constant across all network sizes.
For instance, when we increase the size of the hidden layers in the encoder and decoder from $200$ to $300$ dimensions, the sequence-to-sequence models are able to generalize on datasets where the transition coverage is lower; see \cref{fig:20_acc_dropoff}, where the purple line is the average accuracy of the larger model.
With the exception of the higher transition coverage threshold, these models follow the same trend as their lower dimensional counterparts.\looseness=-1

\paragraph{OSTIA is Slow.} 
In terms of wallclock time, we find that an open-source implementation of OSTIA\footnote{The code is available at  \href{https://github.com/alenaks/OSTIA}{\texttt{github.com/alenaks/OSTIA}}.} does not scale to dataset sizes above $1000$. 
This makes it impossible to perform an apples-to-apples comparison of our neural sequence-to-sequence models against OSTIA.
On the one hand, reducing the size of the training dataset disadvantages OSTIA. 
On the other hand, providing OSTIA with the full $20,000$ samples did not terminate after $3$ days on a single dataset. 
OSTIA provably halts after a finite number of steps, but given the above, a proper comparison with neural models is not possible.\looseness=-1

\begin{figure*}
\centering
\begin{adjustbox}{width=.9\linewidth}
    		\begin{tikzpicture}[->,>=stealth',shorten >=1pt,auto,node distance=3cm,
			semithick, bend angle=20]
			\tikzstyle{every state}=[fill=white,draw=black,text=black]
			\tikzstyle{bluestate}=[state,fill=white, draw=blue, text=black]
			
			\node[state, fill=lightgray!40] (q0)	{\textbf{\textit{$q_0$}}};
			\node[state,fill=lightgray!40] (q1) [right of = q0] {$q_1$};
	        \node[state,fill=lightgray!40, accepting] (q4) [right =6cm of q1] {$q_4$};   
                \node[state, fill=lightgray!40] (q2) [above =2cm of q4] {$q_2$};		
                \node[state, fill=lightgray!40] (q3) [below =2cm of q4] {$q_3$};

			\path
			(q0) edge node {\scriptsize jump / $\varepsilon$} (q1)
			(q1) edge[bend left, above] node[above left] {\scriptsize around / $\varepsilon$} (q2)
			(q1) edge[bend right, above] node[below left] {\scriptsize opposite / $\varepsilon$} (q3)
            
            (q1) edge[bend left =10] node[above] {\scriptsize left / LTURN JUMP} (q4)
            (q1) edge[bend right =10] node[below] {\scriptsize right / RTURN JUMP} (q4)

            (q2) edge[bend left] node [right] {\scriptsize right / RTURN (x4) JUMP} (q4)
            (q2) edge[bend right] node[left] {\scriptsize left / LTURN (x4) JUMP} (q4)

            (q3) edge[bend left] node[left] {\scriptsize left / LTURN (x2) JUMP} (q4)
            (q3) edge[bend right] node[right] {\scriptsize right / RTURN (x2) JUMP} (q4)
			
			;
			
			\path[blue]
			
			;
			\end{tikzpicture} 
	\end{adjustbox}	    

    \caption{A part of the minimal SFST encoding SCAN; in its entirety, it has $> 7000$ states.}
    \label{fig:scan}
   \vspace{-12pt}    
\end{figure*}
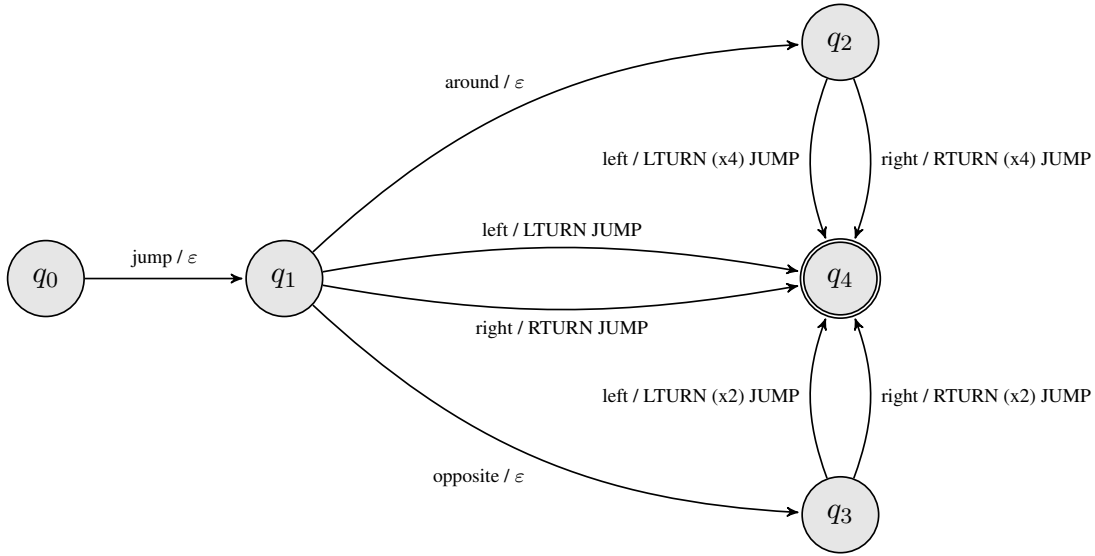

\section{What about SCAN?}\label{sec:discussion}

We now turn to the SCAN dataset and examine it in light of our findings above.
First, we encode SCAN as an SFST.
In so doing, we find it to be an outlier in terms of the number of states it requires, which far exceeds the $100$ states of our largest SFST.
In fact, we calculate that the full SFST encoding SCAN has $7,728$ states; see a small example in \cref{fig:scan}.
With a finite dataset size of $20,000$ input--output pairs, it should not be possible to learn SCAN with high accuracy. 
However, unlike other datasets of a similar size, SCAN turns out to be nearly perfectly learnable in our (random-split) experiments.
% This naturally raises the question of what makes SCAN special.\looseness-1
This result stands in contrast to our randomly generated SFSTs, which exhibit a consistent relationship between the complexity of a formal language and its learnability.\looseness=-1

We offer a possible theoretical behavior for SCAN's surprising learnability.
The class of subregular relations has a native complexity metric built into the formalism: The size of the SFST itself, as measured by the number of transitions. 
However, the results on the SCAN dataset indicate that
the number of transitions in an SFST is \emph{not} the native complexity metric of neural networks under consideration.
This should not come as a surprise because neural networks can learn context-free transductions, which are not even representable by an SFST. 
%Specifically, SCAN is very learnable despite requiring a large number of transitions to encode.
While it is interesting that we empirically identify a scaling law that consistently applies to our randomly sampled SFSTs, it is not the whole story.\looseness=-1

Indeed, SCAN is not a randomly sampled dataset from the class of SFSTs: It is generated by a hand-crafted synchronous context-free grammar \cite[Figure 6]{lake-baroni-2018-generalization}.
While SCAN is a complex SFST, requiring thousands of transitions, it can be encoded by a very small synchronous context-free grammar.
Casting SCAN as an SFST therefore misrepresents its native complexity.
Thus, it is left for future work to identify a class of automata whose native complexity metric can consistently predict the learnability of arbitrary language tasks; such a class will surely be higher than SFSTs on the Chomsky hierarchy given our reported results.\looseness=-1

\section{Related Work}\label{sec:related}
Our paper builds on two common strains of research: The construction of datasets to benchmark compositional behavior in neural networks and 
research in grammatical inference.\looseness=-1

\subsection{Compositionality Datasets}
There is a growing number of artificial language datasets focused on compositionality.
\citet{lake-baroni-2018-generalization} introduced a SCAN dataset, made up of simple navigational text commands. The task is to translate the command in the simple natural language into sequences of actions. One successor to SCAN is the NACS dataset \citep{bastings2018jump}, which is comparable to SCAN, but instead of mapping multiple input signals to a single duplicated output symbol (e.g., \inword{walk twice} $\rightarrow$ \outword{WALK WALK}),  NACS does the opposite (\outword{WALK WALK} $\rightarrow$ \inword{walk twice}).
Since SCAN is a finite language, its inverse NACS is also a finite language, and it can, thus, also be encoded as an SFST.
However, this does not hold true for general SFSTs.
Inverting an SFST often results in non-subsequential transducer because the output tape of SFSTs is, in general, not deterministic.
Another successor is gSCAN \cite{ruis-etal-2020-benchmark} focuses on grounding SCAN-like commands in states of a grid world.
This makes gSCAN closer to \citeposs{mikolov-etal-2016-roadmap} grid world grounding for their agents. 
In contrast to SCAN, gSCAN requires the agent to learn differences between sizes and colours of different geometric shapes and interact with them, by moving them around the grid world.
Executing a gSCAN command is therefore much more difficult than to execute its SCAN counterpart. 
As \citet{ruis-etal-2020-benchmark} assert, the gSCAN dataset removes artefacts in SCAN which are not central to the compositional generalization. They find that models perform worse on gSCAN than on SCAN.
More recently, \citet{bogin} identify that unobserved ``local structures'' in compositionality datasets are harder to learn if no similar structures are observed during training.\looseness-1

\subsection{Grammatical Inference}
Grammatical inference studies the ability to learning classes of formal language from data.
Our work focuses on the learning of a restricted class of functions generated by a correspondingly restricted class of finite-state transducers. 
This allows us to synthesize our study of compositionality in neural models as rule-based inference by neural models, which we can restrict in principled ways.
Finite-state machines generalise many techniques in NLP: probabilistic finite-state automata, hidden Markov models, Markov chains, $n$-grams, probabilistic suffix trees, deterministic stochastic probabilistic automata, weighted automata, and other syntactic objects which generate distributions over sets of possible infinite cardinality of strings, sequences, words, trees, and graphs \citep{VidalFSM}.
Many grammatical inference studies of neural networks test them on samples drawn from some deterministic finite-state acceptors  \citep{cleeremans1989finite}.
See \citet{jacobsson2005rule} for a review.\looseness=-1

Others experiment with neural networks to see if they can learn languages higher up the Chomsky hierarchy.
LSTMs \citep{hochreiter1997long} can perform dynamic counting and variably learn simple counter languages such as some $k$-Dyck languages and $a^{n}b^{n}$ patterns \citep{weiss-etal-2018-practical,suzgun-etal-2019-lstm,bhattamishra-etal-2020-practical, hewitt-etal-2020-rnns}, which are generated by a finite-state machine with a counter on top \citep{schutzenberger1962finite}. In contrast, \citet{Avcu+2017-SCDL-standalone} show that LSTM and other RNN architectures often fail to learn long-distance dependencies drawn from simpler subregular language classes, even on large benchmarks \citep{mahalunkar-kelleher-2019-multi}.
\citet{nelson-etal-2020-probing} study the inference of sequence-to-sequence networks, showing that RNN, LSTM, and GRU \citep{cho2014learning} systematically fail to learn a wide range of regular string copying functions generated from a family of two-way transducers, which characterize regular string-to-string functions. When augmented with attention, they reliably learn every function, and the attention history mirrors the derivations of the corresponding two-way transducers. These independently productive strands of work in compositionality and inference suggest that our work is a reasonable starting point for future interactions.

\section{Conclusion}
We study whether neural sequence-to-sequence models are capable of learning string-to-string transductions with Montague-style compositionality, i.e., where compositional behavior is defined to be homomorphic.
To execute our study, we first provide a theoretical justification of why SFSTs meet Montague's definition.
In the empirical portion of the paper, we randomly sample $2800$ SFSTs using the process described in \cref{sec:sampling-sfsts}, and, then, sample input--output pairs from each SFST to create our unique string-to-string transduction datasets.
We find that neural networks tend to either generalize completely or fail miserably---with little middle ground. 
Moreover, we identify a simple complexity metric, transition coverage, that seems to reliably allow us to predict when an SFST in our randomly sampled dataset is learnable from the dataset sampled from it.\looseness=-1

Finally, our paper discusses how analyzing SCAN as an SFST provides a counterexample to our contention that transtion coverages reliably predicts the learnability of an SFST from a given dataset.
It seems that while transition coverage is a good metric for subregular languages, there are datasets generated by synchronous context-free grammars that are learnable \emph{despite} requiring a large number of transitions when encoded as an SFST.
For example, SCAN's learnability is likely due to the fact that the synchronous context-free grammar used to generate it is relatively small and, thus, under a metric such as production coverage it would be considered simple.
In conclusion, to get a more complete view of the factors underlying learnability, it may be fruitful to consider not only SFSTs, but other formalisms that describe more complex formal relations.
We hypothesize that there might yet be a class of automata whose native complexity will more consistently predict the learnability of a language task.\looseness=-1

\section*{Acknowledgements}
We would like to thank the following people for their valuable insights that made this work possible:
Tim Vieira, Hagen Blix, Dieuwke Hupkes, Douwe Kiela, Brenden Lake, Koustuv Sinha, Anej Svete, Alexandra Butoi, Eleanor Chodroff, Darcey Riley, and William Merrill. 
We also acknowledge Jennifer C. White, who, post publication, spotted some infelicities in the technical exposition which we subsequently corrected. 
\looseness=-1

\bibliography{anthology, custom}
\bibliographystyle{acl_natbib}

\appendix
\onecolumn

\section{Proofs}\label{appendix}
\singlestate
\begin{proof}
\noindent $\Rightarrow$
Assume a homomorphism $f$ and an alphabet $\inputalphabet$ are given.
Construct a finite-state transducer with $\states = F = \{\initialstate\}$.
Let $\outputalphabet = \{f(\sigma) \mid \sigma \in \inputalphabet\}$. 
Create a transition $\initialstate \xrightarrow[ ]{\sigma / f(\sigma)} \initialstate$ for every $\sigma \in \inputalphabet$.
This finite-state transducer is minimal (i.e. non-empty and already has one state), therefore it is the minimal representation of $f$; see \citet{CHOFFRUT2003131} for more details.\looseness=-1

\noindent $\Leftarrow$
Since $\transducer = \transtuple$ is non-empty, we know that its only state $\initialstate$ is also a final state.
Thus, $\transducer$ maps $\varepsilon \mapsto \varepsilon$, i.e., the identity element to the identity element.
Let $\str \in \kleene{\Sigma}$ and suppose that under $\transducer$ we have $\str \mapsto \strout$.
Suppose $\str = \str' \str''$. 
Since $\transducer$ is subsequential, there is a unique path in $\transducer$ that transduces $\str \mapsto \strout$ with exactly $|\str|$ transitions.
Let $\strout'$ be the output yield of the first $|\str'|$ transitions and let $\strout''$ be the output yield of the next $|\str''|$ transitions.
Thus, $\str' \mapsto \strout'$ and 
$\str'' \mapsto \strout''$ which shows that $\transducer$ is a homomorphism since  $\strout = \strout' \strout''$ and $\transducer$ a single-state transducer and we remain in that state.
Finally, we assumed completeness so that $f$ is a total function over $\kleene{\inputalphabet}$.\looseness=-1
\end{proof}

\monoid
\begin{proof}
We define $\boldsymbol{E}$ to be the identity element, i.e., for any $\boldsymbol{A} \in \mathcal{M}^*$, we define  $\boldsymbol{E} \otimes \boldsymbol{A} \defeq \boldsymbol{A} \defeq \boldsymbol{A} \otimes \boldsymbol{E} = \boldsymbol{A}$.
Closure follows from the fact for $\boldsymbol{A}, \boldsymbol{B} \in \mathcal{M}^+$, we have that $\boldsymbol{A} \otimes \boldsymbol{B}$ has by construction exactly one non-zero entry in every row, as elaborated upon in the main text.
To check associativity, consider
\begin{equation}
   \left(\boldsymbol{M}^{\boldsymbol{\sigma}} \otimes \boldsymbol{M}^{\boldsymbol{\sigma}'} \otimes \boldsymbol{M}^{\boldsymbol{\sigma}''}\right)_{ik} = M^{\boldsymbol{\sigma}} _{ij} M^{\boldsymbol{\sigma}'}_{jj'} M^{\boldsymbol{\sigma}''}_{j'k}
\end{equation}
for some $i, j, j', k$.
By the associativity of string concatenation (including when it is augmented to include the zero string), we have that
\begin{equation}
     (M^{\boldsymbol{\sigma}} _{ij} M^{\boldsymbol{\sigma}'}_{jj'}) M^{\boldsymbol{\sigma}''}_{j'k} =  M^{\boldsymbol{\sigma}} _{ij} (M^{\boldsymbol{\sigma}'}_{jj'} M^{\boldsymbol{\sigma}''}_{j'k})
\end{equation}
which in turn implies that
\begin{equation}
   \left((\boldsymbol{M}^{\boldsymbol{\sigma}} \otimes \boldsymbol{M}^{\boldsymbol{\sigma}'}) \otimes \boldsymbol{M}^{\boldsymbol{\sigma}''}\right)_{ik} =    \left(\boldsymbol{M}^{\boldsymbol{\sigma}} \otimes (\boldsymbol{M}^{\boldsymbol{\sigma}'} \otimes \boldsymbol{M}^{\boldsymbol{\sigma}''})\right)_{ik}
\end{equation}
Thus, we have that $(\mathcal{M}^*, \otimes)$ is a monoid.
\end{proof}

\matrix

\begin{proof}
We define $f$ as follows.
First, we define $f(\boldsymbol{E}) \defeq \boldsymbol{E}$.
Next, consider an arbitrary element $\mathcal{M}^*_\inputalphabet$.
Because $\mathcal{M}^*_\inputalphabet$ is a Kleene closure, we may write this arbitrary element 
as $\boldsymbol{M}_{\inputalphabet}^{\sigma_1} \otimes \cdots \otimes \boldsymbol{M}_{\inputalphabet}^{\sigma_K}$.
Now we define $f(\boldsymbol{M}_{\inputalphabet}^{\sigma_1} \otimes \cdots \otimes \boldsymbol{M}_{\inputalphabet}^{\sigma_K}) \defeq \boldsymbol{M}_{\outputalphabet}^{\sigma_1} \otimes \cdots \otimes \boldsymbol{M}_{\outputalphabet}^{\sigma_K}$.
Thus,
\begin{align*}
&f(\boldsymbol{M}_{\inputalphabet}^{\sigma_1} \otimes \cdots \otimes  \boldsymbol{M}_{\inputalphabet}^{\sigma_K})  \\
&= f\left((\boldsymbol{M}_{\inputalphabet}^{\sigma_1} \otimes \cdots \otimes  \boldsymbol{M}_{\inputalphabet}^{\sigma_k}) \otimes (\boldsymbol{M}_{\inputalphabet}^{\sigma_{k+1}} \otimes \cdots \otimes \boldsymbol{M}_{\inputalphabet}^{\sigma_K})\right) \\ &= \left(\boldsymbol{M}_{\outputalphabet}^{\sigma_1} \otimes \cdots \otimes \boldsymbol{M}_{\outputalphabet}^{\sigma_{k}}\right)  \otimes \left(\boldsymbol{M}_{\outputalphabet}^{\sigma_{k+1}} \otimes \cdots \otimes  \boldsymbol{M}_{\outputalphabet}^{\sigma_K}\right)\\
&= \boldsymbol{M}_{\outputalphabet}^{\sigma_1} \otimes \cdots \otimes  \boldsymbol{M}_{\outputalphabet}^{\sigma_K}
\end{align*}
This proves $f$ is a homomorphism.
\end{proof}

\end{document}